\RequirePackage{fix-cm}
\documentclass[twocolumn]{svjour3}          
\smartqed  

\usepackage{amssymb}
\usepackage{graphicx}
\usepackage[ruled,vlined,linesnumbered]{algorithm2e}
\usepackage{float}
\usepackage{color}
\usepackage{soul}
\usepackage{amsmath}
\usepackage{caption}
\usepackage{enumitem} 
\usepackage{makecell}

\usepackage{url}
\begin{document}

\title{Research Issues in Mining User Behavioral Rules for Context-Aware Intelligent Mobile Applications}

\titlerunning{Research Issues in Mining User Behavioral Rules for Context-Aware Intelligent Mobile Applications}        

\author{Iqbal H. Sarker*}


\institute{Department of Computer Science and Software Engineering, Swinburne University of Technology, \\ Melbourne, VIC-3122, Australia. \\ Email: msarker@swin.edu.au}

\date{Received: date / Accepted: date}

\maketitle

\newcommand\blfootnote[1]{%
	\begingroup
	\renewcommand\thefootnote{}\footnote{#1}%
	\addtocounter{footnote}{-1}%
	\endgroup
}

\makeatletter
\def\footnoterule{\relax%
	\kern 0pt
	\hbox to \columnwidth{\hfill\vrule width .95\linewidth height 0.6pt\hfill}
	\kern .1pt}
\makeatother

\blfootnote{(Preprint submitted to a Springer journal)}

\begin{abstract}
Context-awareness in smart mobile applications is a growing area of study, because of it's intelligence in the applications. In order to build context-aware intelligent applications, mining \textit{contextual behavioral rules} of individual smartphone users utilizing their phone log data is the key. However, to mine these rules, a number of \textit{issues}, such as the quality of smartphone data, understanding the relevancy of contexts, discretization of continuous contextual data, discovery of useful behavioral rules of individuals and their ordering, knowledge-based interactive post-mining for semantic understanding, and dynamic updating and management of rules according to their present behavior, are investigated. In this paper, we briefly discuss these \textit{issues} and their \textit{potential solution} directions for mining individuals' behavioral rules, for the purpose of building various context-aware intelligent mobile applications. We also summarize a number of real-life \textit{rule-based applications} that intelligently assist individual smartphone users according to their behavioral rules in their daily activities.

\end{abstract}

\section{Introduction}
Nowadays, mobile phones have become one of the primary ways in which people around the globe communicate with each other. According to \cite{number-of-mobile-phone-users}, cellular network coverage has reached 96.8\% of the world population and this number even reaches 100\% of the population in developed countries. These devices, particularly the smart mobile phones have transformed over a period of time from merely communication tools to smart and highly personal devices enabling to assist the users in their variety of day-to-day situations in their daily life. 

In the real word, users' interest on ``Mobile Phones'' is more
and more than other platforms like ``Desktop Computer''
or ``Tablet Computer'' over time \cite{sarker2018MobileDataScience}. People use mobile phones not only for voice communication between individuals but also for various activities such as applications (mobile apps) using, Internet browsing, e-mailing, using online social network, instant messaging etc \cite{pejovic2014interruptme}. Recent advances in the sensing capabilities of smart mobile phones make them enable to collect the rich contextual information and users' various activity records with mobile phones through the device logs. These historical mobile phone data are simply as the collection of the past contexts and user's activities with the mobile phones for these past contexts. These are phone call logs \cite{sarker2016phone} having phone call activities, app usages logs \cite{srinivasan2014mobileminer} having various mobile application usages, mobile phone notification logs \cite{mehrotra2016prefminer} having the responses with various notifications from different applications, web logs \cite{halvey2005time} having Internet browsing activities of the mobile phone users. The main characteristic of such kind of phone log data is that it contains the actual diverse activities of the users in different contexts in their real world life.

Modeling smartphone user behaviors by developing various computational machine learning methods (rule-based learning) in order to analyze different behavioral patterns in different contexts, and eventually predict the next behaviors or detect strange behaviors utilizing such mobile phone data, can be used for building various real-world intelligent systems. For instance, smart context-aware mobile communication system, intelligent notification management system, context-aware mobile recommender system, and various rule-based predictive systems, can be built using the extracted behavioral rules or patterns, in order to assist them intelligently according to their personal needs. In the later part of this paper, we briefly discuss about these applications based on the extracted behavioral rules of individual mobile phone users, in which we are interested in.

Context-awareness in such mobile applications is the core part to make these applications intelligent \cite{sarker2018MobileDataScience}. Based on the contextual behavioral rules of individuals, the applications can behave intelligently according to the users' current situation and preferences. Thus, mining individuals' \textit{contextual behavioral rules} from their phone log data, is a key research area, in order to build such intelligent applications. However, mobile phone users' behaviors are not identical to all in the real world. Individual user may behave differently with their phones in different contexts according to their day-to-day situations. Let's consider an example of a smart phone call handling service, a mobile phone user typically `declines' the incoming phone calls, if she is in a `meeting'; however, she `answers' the incoming call if the call comes from her `boss' as it seems to be significant for her. Hence, [decline, answer] are the user phone call behaviors, and [meeting, boss] are the associated social contexts, i.e., meeting represents the social activity or situation, and boss represents the social relationship of that user. Similarly, other relevant contexts such as temporal, or spatial context can play a significant role in her diverse activities with mobile phones in her real world life. According to Dey et al. \cite{dey2001understanding}, context is defined as ``any information that can be used to characterize the situation of an entity''. In our analysis, an individual smartphone user is represented as an entity, and temporal, spatial or social contexts are taken into account as the associated multi-dimensional contexts of the users for the relevant context-aware intelligent applications, highlighted above. 

In this paper, we mainly formulate the problem of mining behavioral rules of individual mobile phone users based on these contexts, utilizing their smartphone data. To mine these rules, we investigate a number of \textit{issues}, such as the quality of smartphone data, understanding the relevancy of contexts, discretization of continuous contextual data, discovery of useful behavioral rules and their ordering, knowledge-based interactive post-mining, and dynamic updating and management of rules according to their present behavior. In this paper, we briefly discuss these \textit{issues} and their \textit{potential solution} directions for mining the contextual behavioral rules of individual smartphone users. The extracted behavioral rules of individuals can be used to build various context-aware intelligent systems, in order to provide them the personalized services according to their needs, in a context-aware pervasive computing environment.

In particular, the contributions of this paper are summarized as below: 

\begin{itemize}
	\item We highlight the importance of contextual behavioral rules of individual mobile phone users for the purpose of building various context-aware intelligent applications.
	
	\item We briefly discuss a number of research issues for mining user contextual behavioral rules from smartphone data and corresponding potential solution directions to overcome these issues. 
	
	\item We explore a number of potential rule-based mobile applications for the end users based on their contextual behavioral rules to assist them intelligently in their daily activities.
\end{itemize}

The rest of the paper is organized as follows. We discuss about the contextual behavioral rules of individual mobile phone users and the features of such rules in Section \ref{Features of Contextual Behavioral Rules}. In Section \ref{Research Gap}, we review the most recent works related to context-aware intelligent systems based on user behavioral rules. In Section \ref{Challenges and Opportunities}, we briefly discuss about the investigated research issues for mining contextual behavioral rules of individuals and their potential solution directions to overcome these issues. A number of potential real-life applications for individual mobile phone users based on their contextual behavioral rules are explored in Section \ref{Applications}. Finally, Section \ref{Conclusion} concludes this paper.

\section{Contextual Behavioral Rules}
\label{Features of Contextual Behavioral Rules}
A rule $(A \Rightarrow C)$ is any statement that relates two principal components, the rule's left-hand-side (antecedent, A) and the rule's right-hand-side (consequent, C) together. An antecedent states the condition (IF) at issue and a consequent states the result (THEN) held from the realization of this condition, i.e., (IF-THEN logical statement). According to this definition, we define a \textit{contextual behavioral rule} as $[contexts \Rightarrow behavior]$, where the $contexts$ (antecedent) represents the contextual information (one or many) of an individual mobile phone user, and the $behavior$ (consequent) represents his/her mobile phone usages behavior for that contexts. An example of a contextual behavioral rule of an individual mobile phone user would be ``if a user is in a meeting (context), she declines (behavior) the incoming phone calls'', and represented in the rule format as $[meeting \Rightarrow decline]$. However, ``if the calls come from her mother (another context), she answers (behavior) the incoming calls'' and represented in the rule format as $[meeting, mother \Rightarrow answer]$, A set of discovered such contextual behavioral rules of individual mobile phone users utilizing their own phone log data, provide a basis on which the personalized services can be provided in various real life mobile applications to assist them intelligently in their daily activities, in a context-aware pervasive computing environment.

In order to mine \textit{contextual behavioral rules} of individual mobile phone users utilizing their smart mobile phone data, the following features or properties are needed to take into account. These are:

\begin{enumerate}[label=(\roman*)]
	\item {\bf Context-Aware}: Real-world phone log data usually comprises a set of features whose interpretation depends on some \textit{contextual information}. Context consists of any circumstantial factors users are involved in, such as temporal context - time-of-the-day (24-hours), days-of-the-week (Monday, Tuesday, ..., Sunday), spatial context - user's current location (e.g., office), and social context - user's social activity or situation (e.g., meeting), user's social relationship between individuals (e.g., mother) etc. Such contextual features and related patterns are of high interest to be discovered from the mobile phone data and analyzed in order to obtain the right meaning in behavioral rules. Therefore, the behavioral rules of individual mobile phone users should be \textit{contextual}, which refers to the search for associations between these contexts such that the strength of their implication depends on a contextual feature. \\ 
	
	\item {\bf Behavior-Oriented}: In the real world, mobile phone usages behavior of individuals are not identical to all, may differ from user-to-user in different contexts (e.g., in a meeting). For instance, a mobile phone user (say, an employee, Alice) typically `declines' the incoming phone calls in a meeting, on the other hand, another individual (say, her boss) may `answer' the incoming phone call during that meeting. For the similar context (e.g., $social \; situation \rightarrow meeting$), different individuals (Alice and her boss) may behave differently with their own cell phones in the real world. Thus, the rules should represent individual's unique behavioral patterns rather than identifying the best decision for a particular contextual information. According to Ross \cite{decisionRule-VS-BehavioralRule}, ``Unlike decision rules, behavioral rules do not pertain directly to determining the best or most appropriate answer (outcome) among alternatives''. Therefore, the discovered contextual rules of individual mobile phone users should be individual's \textit{behavior-oriented} according to their unique patterns in the dataset. \\
	
	\item {\bf Individual's Preference}: Consistency in mobile phone usages behaviors of individuals determines the strength of their behavioral patterns and corresponding rules. However, mobile phone users' behaviors are not consistent in the real world. For instance, one individual always (say, 100\%) `declines' the incoming phone calls when s/he is in office (`\textit{Consistent behavior}). Another individual may `declines' most of the incoming phone calls (say, 85\%), `answers' (say, 10\%), and `misses' (say, 5\%) when in the office (`\textit{Inconsistent behavior}). Thus, the strength of the behavioral rule for a particular context (e.g., $location \rightarrow office$) may not be similar to all individuals. Therefore, the behavioral rules should be based on \textit{individual's preference} that selects the minimum `rule strength' (i.e., known as confidence) valid for that particular user, which differ from user-to-user according to their own interests. Such decision making is based on the past behavioral evidence (not a random process) of individuals for a particular confidence level preferred by them.
\end{enumerate} 

Example 1. Let's consider a phone call handling service of a smart phone user, Alice. She typically `declines' most of the incoming phone calls (83\%), if she is in a `meeting' (social activity or situation); however, she always (100\%) `answers' the incoming phone calls in that meeting, if the call comes from her `boss' (social relationship). Then the following contextual rules would represent Alice's phone call response behavior for the associated contexts. Say, the rules are discovered for a particular confidence threshold (80\%) preferred by Alice, utilizing her smart phone call log data: 
\begin{equation*}
\begin{split}
& (i) Meeting \Rightarrow Decline \; (Conf = 83\%)\\
& (ii) Meeting, Boss \Rightarrow Answer \; (Conf = 100\%)\\
\end{split}
\end{equation*}

Hence, [Meeting, Boss] are the contextual information that have a strong influence on Alice to make her call handling decisions during meeting, [Decline, Answer] are Alice's phone call response behaviors depending on that contexts, and the [confidence threshold (80\%)] for discovering these phone call response behavioral rules is preferred by Alice, i.e., the rules that satisfy this confidence threshold ($\ge$80\%) are discovered for the user Alice, which can be used to build a context-aware smart phone call interruption management system for the user Alice, that intelligently assists her by handling the incoming phone calls according to her current contextual information. \\

Example 2. Let's consider a mobile application management service of another smart phone user, Bob. He typically uses `Facebook' (92\%) between 09:00AM and 10:00AM (temporal); however, in most cases he uses `Microsoft Outlook' (98\%), if he is in `office' (location) at that time period. Then the following contextual rules would represent Bob's mobile apps usages behavior for the associated contexts. Say, the rules are discovered for a particular confidence threshold (90\%) preferred by Bob, utilizing his smart phone apps usage log data: 

\begin{equation*}
\begin{split}
& (i) [09:00AM-10:00AM] \Rightarrow Facebook \; (92\%)\\
& (ii) [09:00AM-10:00AM], Office \Rightarrow Outlook \; (98\%)\\
\end{split}
\end{equation*}

Hence, [09:00AM-10:00AM, Office] are the contextual information that have a strong influence on the user Bob to make his app usages decisions between 09:00AM and 10:00AM, [use Facebook, use Microsoft Outlook] are Bob's mobile apps usages behaviors depending on that contexts, and the [confidence threshold (90\%)] for discovering these app usages behavioral rules is preferred by Bob, i.e., the rules that satisfy this confidence threshold ($\ge$90\%) are discovered for the user Bob, which can be used to build a smart mobile app management system for the user Bob, that intelligently assists him by predicting his future usages according to his current contextual information.

\section{Existing Research}
\label{Research Gap}
In the area of mining mobile phone data, a significant amount of research has been done. For instance, a number of works \cite{rawassizadeh2016scalable,mukherji2014adding,bayir2010web,paireekreng2009time,jayarajah2014exploring,do2010their,zulkernain2010mobile,ma2012habit,khalil2005improving,dekel2009minimizing,phithakkitnukoon2010activity,mehrotra2016prefminer,srinivasan2014mobileminer,zhu2014mining} based on contextual information or rules utilizing mobile phone data has been done for various purposes. Hence, we briefly review the most recent relevant works based on contextual behavioral rules of individual mobile phone users. 

For instance, in \cite{mehrotra2016prefminer}, Mehrotra et al. proposed an approach PrefMiner for mining user's preferences for intelligent mobile notification management. The design of preference miner is based on mobile notification dataset. They learn individual users' preferences for receiving notifications based on automatic extraction of association rules by mining their interaction with mobile phones utilizing such notification dataset. Based on these rules, they build an interruption management system that handles the mobile phone notifications according to the preference of individuals. In another work \cite{srinivasan2014mobileminer}, Srinivasan et al. proposed an approach MobileMiner for mining the frequent patterns on the mobile phones. In their work, they mine the mobile phone data related to mobile apps and phone calls. They extract the interesting behavioral patterns based on different contexts and represented as association rules for each individual user. Their work ranges from calling activity patterns to place visitation patterns of individuals. In their analysis, they finally showed how the extracted association rules or patterns can be used to improve in launching apps or calling contacts, by predicting the next app or contact according to the extracted association rules. In another recent work, Zhu et al. \cite{zhu2014mining} proposed an approach ContextLogsMiner for mining mobile user preferences for personalized context-aware recommendation. They take into account both context independent and context dependent assumptions for mining the context-aware preferences. In their work, they extract the personal context-aware preferences from individuals context logs and represented as rules. Based on these rules they build personalized context-aware recommender systems for the end mobile phone users.

These approaches produce behavioral rules of individuals based on different contexts utilizing their mobile phone data. However, issues are investigated in different aspects while mining such rules from the mobile phone data. In next section, we briefly discuss these \textit{issues} and their \textit{potential solution} directions for mining the contextual behavioral rules of individual smartphone users.

\section{Summarizing Research Issues}
\label{Challenges and Opportunities}
In this section, we summarize a number of research issues related to \textit{mining contextual behavioral rules} of individual mobile phone users. These include the quality of smartphone data, understanding the relevancy of contexts to provide dynamic services, discretization of continuous contextual data as the basis for knowledge discovery, user behavioral rule discovery and ordering, knowledge based interactive post-mining for semantic understanding, and dynamic updating and management of rules according to their present behavior. In the following, we briefly discuss these issues one by one.

\subsection{The Quality of Smartphone Data}
Mobile phone data may contain noise (wrong and/or redundant samples) because of using a variety of sensors or data sources while collecting and storing the data. Such inconsistency in mobile phone dataset is also called \textit{noise}. Simply noise is anything that obscures the relationship between the features or contexts of an instance and it's behavior class in the datasets \cite{frenay2014classification}. The presence of such noisy instances in mobile phone data is a fundamental issue for modeling user behavior with many potential negative consequences. For instance, the over-fitting problem may arise and thereby the prediction accuracy may decrease and the complexity of the machine learning techniques may increase due to the number of wrong or redundant training samples \cite{sarker2017improved}. 

Getting higher classification or prediction accuracy of user behavior by analyzing individuals' mobile phone log data using machine learning techniques (e.g., decision tree) is challenging, as the prediction model requires a training data set free from noise. According to \cite{sarker2017improved}, the effects of noisy instances in the real-life mobile phone data for predicting user behavior as follows:

\begin{itemize}
	\item Noise may create additional behavioral rules that are not interested to individual mobile phone users and make the rule-set unnecessarily larger.
	\item The number of training samples may increase in the dataset and as a result the complexity of the corresponding machine learning based behavioral model for individuals may increase.
	\item The presence of noisy training instances in the dataset may cause over-fitting problem when applying the decision tree classification technique or building a tree-based model, and thus decrease the classification or prediction accuracy of the inferred behavioral model for individual mobile phone users.
\end{itemize}

Thus, identification and elimination of the noisy instances from a training dataset are required to ensure the quality of the training data. A machine learning technique, e.g., naive Bayes classifier based noise detection approach \cite{sarker2017improved} could be useful to identify the inconsistency in the data sets. After removing such inconsistency in the dataset, the remaining noise-free instances could be a quality source of data for mining user behavioral rules for getting higher performance. According to \cite{zhu2004class}, the performance of the machine learning technique based model depends on two significant factors: (i) the quality of the training data, and (ii) the competence of the machine learning algorithm. Therefore, a noise reduction process is required before constructing the model in order to achieve better prediction accuracy of the model based on real life mobile phone data. 

\subsection{Understanding the Relevancy of Contexts}
To realize the need for \textit{contexts} is an important step towards using them effectively in mining contextual behavioral rules of individual mobile phone users. In order to effective use of contexts in the behavioral rules of individual mobile phone users, we need a clear understanding of what contexts have the influence on users to make decisions in different situations. As we aim to discover the contextual behavioral rules of individuals utilizing their mobile phone data, the contexts related to the user are the most relevant. Table \ref{tab:context-examples} shows an example of user's contexts having influence on individuals for making decisions in different situations. However, the \textit{relevancy} of the contexts is application specific, i.e., may vary from one application to another application in the real world.

\begin{table*}
	\begin{center}
		\caption{Various types of user contexts}
		
		\label{tab:context-examples}
		\begin{tabular}{l|c} 
			\textbf{Context Category} & \textbf{Context Examples}\\
			\hline
			Temporal Context & \makecell{User's activity occuring date (YYYY-MM-DD), \\ time (hh:mm:ss), period (e.g., 1 hour, 10:00am-12:00pm), \\ weekday (e.g., Monday), weekend (e.g., Saturday), etc.}\\
			\hline
			
			Spatial Context & \makecell{GPS based continuous locations or \\ corresponding coarse level of locations such as \\ office, work, home, market, restaurant, \\ vehicle, playground etc.}\\
			\hline
			
			Social Context & \makecell{User's social activity or situation \\ such as professional meeting, lecture, seminar, \\ lunch break, dinner, etc., \\ and/or, social relational context \\ such as family, friend, professional or work relationship, \\ significant one, unknown, etc.}\\
			\hline
		\end{tabular}
	\end{center}
\end{table*}

Let's consider a \textit{personalized smart mobile app management system} that can predict individual's future application usages (e.g., Skype, Whatsapp, Facebook, Gmail, Microsoft Outlook, etc.) according to his/her contextual information. Say, on Weekdays between 09:00AM and 10:00AM, the user typically uses `Microsoft Outlook' for mailing purposes, when she is in her office. In order to search this particular mobile application among a huge number of installed apps in her mobile phone, the user's contexts such as temporal (in Weekdays between 09:00AM and 10:00AM), and location (at office), might be relevant to intelligently assist herself. Let's consider another example, a \textit{smart phone call interruption management system}, where more contexts might be relevant. In the real-world, the mobile phones are considered to be `always on, always connected' device but the mobile users are not always attentive and responsive to incoming communication \cite{chang2015investigating}. Say, on Monday between 09:00AM and 11:00AM, a user attends a regular meeting in her office. Typically, she declines the incoming phone calls during that time period as she does not want to be interrupted with phone calls during the meeting. However, if the phone call comes from her boss or her mother, she wants to answer the call as it seems to be significant for her. According to this example, user's phone call response behaviors are not only related to the above contexts, location (e.g., at office), and temporal (e.g., on Monday, between 09:00AM and 11:00AM), but also related to the additional contexts, social situation (e.g., in a meeting), and social relational context (e.g., boss or mother). 

According to the above real-world examples, it is clear that the relevancy of user's contexts varies from application to application in the real world. Thus, a better understanding of context relevancy according to individual users' need, will help mobile application developers to choose what context to use in their applications in order to provide the personalized services to assist them intelligently in their daily activities.

\subsection{Discretization of Continuous Contextual Data}
Discretization is one of the most important preprocessing techniques used in the area of data mining, to be used as the basis for discovering the useful knowledge or rules. The discretization process mainly transforms the continuous numerical attribute values into the discrete or nominal attribute values based on some conditions. In other words, it transforms the quantitative data into qualitative data, with a finite number of intervals, obtaining a non-overlapping partition in a continuous domain, such as time, location. The nature of such continuous data includes: may large in data size, contains high dimensionality, and to update continuously. Assuming a data set consisting of $N$ samples and $C$ target classes, a discretization algorithm would discretize the continuous attribute $A$ in this data set into $m$ discrete intervals, $Dis=${$[d_0, d_1]$, $[d_1, d_2]$,..., $[d_{m-1}, d_m]$}, where $d_0$ represents the minimal value, $d_m$ represents the maximal value, and $d_i < d_{i+1}$, for $i = 0,1,..,m-1$. Such a discrete result $Dis$ is called a discretization scheme on attribute $A$, and $<p={d_1,d_2,..,d_{m-1}}>$ is the set of cut points of attribute $A$. 

Time-of-the-week is the most important continuous context that impacts on user behavior in a mobile-Internet portal \cite{halvey2005time}. The mobile phones record the exact temporal information (e.g., YYYY-MM-DD hh:mm:ss) of different activities of users' with their mobile phones. Such temporal information is known as the ``time series data", which is numeric and continuous \cite{sarker2017individualized}. However, human understanding of time is not precise in behavior modeling, unlike digital systems. There is a need of time interval, e.g., five minutes, in every routine behaviors. For instance, a college student regularly makes a phone call to her mother in the evening to discuss about her day long studies. It is unlikely that she will call her mother everyday exactly at 6:00PM; she could call one day at 6:14PM and another day at 5:54PM. Thus, a time segment or interval, such as between 5:50PM and 6:15PM, rather than an exact temporal information, is very informative to capture her activity patterns.

To generate such time segments capturing similar behavioral characteristics, an optimal segmentation technique is needed. To mine mobile user behavior, a number of researchers \cite{mehrotra2016prefminer,zhu2014mining,srinivasan2014mobileminer} use large interval based segmentation (e.g., morning[6:00AM-12:00PM]). However, such large segments may not be suitable to produce the meaningful behavioral rules of individuals. The reason is that such large segments may not be able to differentiate individual's diverse activities in a particular segment \cite{sarker2017individualized}. On the other hand, a number of researchers \cite{ozer2016predicting,phithakkitnukoon2010activity} use small interval based segmentation (e.g., 15 minutes)  by taking into account the frequent variations of individual's behaviors. However, in many cases, these small interval based time segments may not be suitable to produce meaningful behavioral rules in terms of support value \cite{sarker2017individualized}.

The main drawback of existing approaches discussed above, is that segmentation used in the applications are not individual's behavior-oriented. Thus, the key requirement is to extract similar behavioral time segments for various days-of-the-week enabling the generation of high confidence rules. A bottom-up segmentation approach \cite{sarker2017individualized} by generating the dynamic temporal segments based on the similar dominant behavioral characteristics in various days-of-the-week, could be useful to get the optimal time periods for the purpose of mining user behavioral rules. In addition to the temporal context, other types of contexts, such as users' location measured by GPS have numerical values that are continuous. Thus, like the discretization of temporal information, a method is needed to be developed to pre-process such continuous location context values to convert them into nominal values before applying the rule mining technique.

\subsection{User Behavioral Rule Discovery and Ordering}
In the area of mobile data mining, association rule learning technique (e.g., Apriori) \cite{agrawal1994fast}, and classification rule learning technique (e.g., Decision tree) \cite{quinlan1993}, are the most common approaches to discover rules of mobile phone users. However, the decision tree-based rules mostly have low reliability \cite{mehrotra2016prefminer}. According to \cite{freitas2000understanding}, it cannot ensure that a discovered classification rule will have a high predictive accuracy due to \textit{over-fitting problem} (do not generalize well from the training data) and \textit{inductive bias} (favoring one hypothesis over another, other than strict consistency with the data being mined). Moreover, it takes into account rigid decision \cite{quinlan1993} as it has no \textit{flexibility} to set user preference (e.g., confidence level) that may vary from user-to-user.

On the other hand, association rule learning technique \cite{agrawal1994fast} is well defined in terms of rule's reliability and flexibility in decisions as it has the own parameter support and confidence \cite{freitas2000understanding}. This technique implicitly assumes all the contexts in the datasets have the same nature, impact and/or similar frequencies and takes into account all the combinations of contexts while producing rules. As a result, it is unable to differentiate between \textit{useful rules} and \textit{useless (redundant) rules}. According to \cite{fournier2012mining}, association rule learning technique produces up to 83\% redundant rules that makes the rule-set unnecessary larger. Therefore, it is very difficult for the decision makers to determine the most interesting ones and consequently makes the decision making process ineffective and more complex. In addition, it is not applicable to provide the real-time, proactive and personalized services as it takes huge amount of training time for mining rules. Srinivasan et al. \cite{srinivasan2014mobileminer} observe a high running time spanning several hours when the association rule mining algorithm Apriori \cite{agrawal1994fast} is applied to mobile context data through their experimental study. Thus, a concise set of \textit{non-redundant association rules} that represent users' behavior in various context dimensions, is needed to discover.

Rather than using the traditional association rule learning technique discussed above, an AGT (Association Generation Tree) \cite{sarker2018mining} based rule learning technique could be useful to generate a set of non-redundant behavioral association rules based on multi-dimensional contexts. As AGT generates both the general and specific rules, the ordering of these rules is also important for the execution of rules according to their priorities. Thus, to take into account the highest number of matched contexts in the rule's antecedent, i.e., specific-to-general, could be useful while ordering the discovered rules.

\subsection{Knowledge-Based Interactive Postmining}
Knowledge-based postmining of the discovered rules, might be another research issue, in terms of semantically generalization of rules, in order to avoid categorical data sparseness, and to make the rules more useful and interesting in a particular domain.

The semantic generalization using the relevant domain knowledge further generalizes the rules according to their semantic relationships. The semantic generalization of rules plays an important role in improving context-based adaptation and in ensuring the correct behavior of individuals in context-aware applications and services. Typically, it is difficult for decision makers to process, interpret and utilize the produced data specific rules in decision making process for many applications. Moreover, to select the best rules according to the query context, fewer rules but closely relevant is important. Thus for effective use of the produced rules, two issues are needed to take into account; (i) discover the fewer closely relevant rules, and (ii) handling categorical data sparseness while applying these rules. The idea of knowledge based multi-level generalization of produced rules can play a role to overcome the above issues.

A semantically generalized association rule mining is at a higher level of abstraction of the discovered rules based on some concepts. Like the regular association rule, a generalized association rule is an implication of the form X$\Rightarrow$Y, where X represents the antecedent and Y represents the consequent of the rule. Such generalized rules also have the parameters support and confidence. If the preferences for this support and confidence are set low, too many rules satisfy these thresholds typically and huge number of corresponding rules are generated. Generalized association rule can help to reduce the search space and combine a number of low support rules into a less number of high support rules by taking into account the use of a \textit{concept hierarchy} for a particular domain.

Let's consider a phone call example in terms of support values of rules. The rules ``incoming phone calls from mother are answered at office'' and ``calls from father are also answered at office'' are generated from the dataset, but do not have the minimum support value. As a result, the rule learning algorithm ignores such rules and causes information losing. The knowledge-based generalized association-rule-learning algorithm extends these baseline rules by aiming at descriptions at the appropriate taxonomy level $-$ for example, ``calls from parents are answered at office'', rather than ``calls from mother are answered at office'' and ``calls from father are answered at office'' separately. Such generalized rule ``calls from parent are answered at office'' using the concept of family relationship, may have the minimum support as this knowledge based generalized rule combines the two separate rules mentioned above and their support values as well. 

Another task is to handle categorical data sparseness with respect to the query context. Let's consider an example, assume the concepts R1, R2, and R3 are semantically related with each other. For a given contexts the produced rules, e.g., if X$\Rightarrow$R1 or R2, then Y$\Rightarrow$A are item-specific, and no rules are proceeded for R3 because of data sparseness. However, if the query context is related to R3, then the system using the above item-specific rule is unable to make a decision. Since R1, R2 and R3 are semantically related, the knowledge-based generalization of these concepts, e.g., if X$\Rightarrow$R1 or R2 or R3, then Y$\Rightarrow$A that is more interesting rule and also applicable for the concept R3. 

Ontology-based approach could be a possible way to use such concepts for a particular domain while generalizing the rules. In general, ontology is an explicit specification of conceptualization and a formal way to define the semantics of knowledge and data. The formal structure of ontology makes it a nature way to encode domain knowledge for the data mining use. According to \cite{maedche2001ontology},  formally, an ontology is represented as $\{O = C, R, I, H, A\}$, where $\{C = C_1, C_2,...,C_n\}$ represents a set of concepts,
and $\{R = R_1, R_2,...,R_m\}$ represents a set of relations defined over the concepts. $I$ represents a set of instances of concepts, and $H$ represents a Directed Acyclic Graph (DAG) defined by the subsumption relation between concepts, and $A$ represents a set of axioms bringing additional constraints on the ontology.

By defining shared and common domain theories, ontologies help people and machines to communicate concisely by supporting semantic knowledge for a particular domain. The reason is that ontology based rule generalization discovers a high level view of the produced rules which as a result produces fewer, but more closely associated rules that becomes more interesting to the decision makers. Because, the rule interestingness strongly depends on user knowledge and goals. The more the knowledge is represented in a flexible, expressive and accurate formalism, the more the rule selection will be effective. So a method is needed that can help to discover the generalized association rules in multiple abstraction levels by taking into account the issues discussed above in the step of post-processing knowledge. 

\subsection{Dynamic Updating and Management of Rules}
Mobile phone log data is not static as it is progressively added to day-by-day according to individual's present behaviors with mobile phones. Since individual's behavior changes over time, the most \textit{recent patterns} (e.g., recency) are more likely to be interesting and significant than older ones for predicting individual's future behavior in a particular contexts. Although, the produced rules (discussed above) are able to predict individuals behavior (long-term prediction based on whole log data), the recency based rules might be more significant for \textit{short-term prediction} in terms of accuracy. Thus, the updating of discovered rules based on individual's recent behavioral patterns and their dynamic management, becomes another challenge as the updates may not only invalidate some existing rules but also make other rules relevant.

A number of research \cite{lee2010adaptive,phithakkitnukoon2011behavior} have considered this issue by taking into account the behavioral patterns of recent mobile phone log data rather than the patterns derived from the entire historical logs. However, the main limitation of existing approaches is that they used a static period of time that is not able to identify the changes in individuals' behavioral patterns. Thus, identifying such changes dynamically and to update the existing rules according to the present behavior of individuals is another research issue for mining smartphone user behavioral rules based on contexts. A recency-based rule updating algorithm that identifies and removes the `outdated rules' (do not represent the present behavior of an individual) \cite{sarker2017understanding} from the existing-rule set, could be useful. These updated behavioral rules could be more effective for using in various intelligent context-aware systems in order to provide personalized services to assist them in their daily activities.

\section{Rule-based Intelligent Mobile Applications}
\label{Applications}
A rule-based system represents knowledge in terms of a set of IF-THEN rules that tells what to do or what to conclude in different situations \cite{grosan2011rule} and can act as a software agent. The target applications of this research are those context-aware personalized applications that have been studied widely in the past few years: smart context-aware mobile communication, intelligent mobile notification management, context-aware mobile recommendation, and various predictive services etc.

\begin{itemize}
	\item Smart Context-Aware Mobile Communication: Mobile phones are considered to be `always on, always connected' device but the mobile users are not always attentive and responsive to incoming communication \cite{chang2015investigating}. As a result, people are often interrupted by the incoming phone calls in various day-to-day situations in their daily life, which not only create disturbance for the phone users but also create disturbance for the people nearby. Such kind of interruptions may create embarrassing situation not only in an official environment, e.g., meeting, lecture etc. but also affect in other activities like examining patients by a doctor or driving a vehicle etc. Sometimes these kind of interruptions may reduce worker performance, increased errors and stress in a working environment \cite{pejovic2014interruptme}. In the real world, calling activity records in device logs (e.g., phone call logs) are a rich resource for mining contextual behavioral rules of individual mobile phone users, for building smart call interruption handling systems, in order to handle the incoming calls intelligently according to their preferences \cite{sarker2016phone}. For instance, in \cite{sarker2018Unavailability}, Sarker et al. present SilentPhone, which is a temporal context based approach to dynamically change the phone ringer mode. Another real-life application could be a smart call reminder system that intelligently searches the desirable contact from the large contact list and reminds a user to make a phone call to a particular person in a particular contextual situation, according to the discovered behavioral rules of individuals based on the user's past calling history. As these applications are personalized and contact specific, data-centric social context \cite{sarker2018DataCentricSocialContext}, can play an important role to build data-driven intelligent systems for the end mobile phone users. \\
	
	\item Intelligent Mobile Notification Management: Now-a-days, a variety of smart mobile applications are available on the app stores. These applications enable the mobile phone users to subscribe too many information channels and actively receive numerous information through such notifications \cite{mehrotra2016prefminer}. However, the mobile phone users do not accept all of such notifications depending on the content type and the sender of the messages \cite{mehrotra2015designing}. According to \cite{mehrotra2016prefminer}, users mostly dismiss notifications that are not useful or relevant to their interests. For examples, in most cases, the notifications of inviting games on social networks, social or promotional emails are swiped away without clicking as having no interest on these notifications. Moreover, the predictive suggestions by various mobile phone applications, e.g., Twitter, Facebook, Linkedin, WhatsApp, Viber, Skype, Youtube, etc. may not be interested to a particular user \cite{mehrotra2016prefminer,kanjo2017notimind,turner2015push}. The reason is that the mobile phone users might get irritated for such uninterested phone notifications. Consequently, in some cases, the users uninstall the corresponding applications from their smart mobile phones in order to avoid receiving such notifications. Individual's behavioral rules based on user's contextual information, might be able to manage such notifications intelligently. For examples, one individual always dismisses promotional email notifications; she accepts birthday reminder notifications of Facebook mostly at night, when she is at home; does not accept Viber or Whatsapp notifications from unknown persons at office. As the behaviors of different individuals are not identical in the real word, such behavioral rules may differ from user-to-user according to their own interests in different contexts. \\
	
	\item Context-Aware Mobile Recommendation: In general, the traditional recommender systems mainly focus on recommending the most relevant items to users among a huge number of items \cite{bobadilla2013recommender}. These recommendation systems do not consider the contextual information while making the recommendation to a particular user \cite{shin2009context}. However, without taking into account such contextual information, it may not be sufficient in making recommendation in some real life scenario. For instance, the recommendation outcome of a travel recommender system in the summer can be well different from the outcome in the winter for a particular user, which depends on the temporal context. Similarly another context, e.g., location information, might have the influence to make different recommendations for the users \cite{park2007location,zheng2010collaborative}. Mobile app recommendation is one of the major part in mobile recommendation \cite{liu2015personalized}. With the rapid development and adoption of mobile platforms such as smartphones and tablets, they have become one of the most important media for social entertainment and information acquisition \cite{zhu2014mining}. In fact, various contexts and corresponding app usages (e.g., Multimedia, Facebook, Gmail, Youtube, Skype, Game, etc.) data is recorded in context-rich device logs which can be used for mining the contextual behavioral rules of individual mobile phone users, i.e., which app is preferred by a particular user under a certain context. Particularly, mining such preferences is a fundamental work for understanding the app usages behaviors of mobile phone users. The extracted behavioral rules utilizing context-logs can be used to provide personalized context-aware recommendation of different mobile phone apps for the mobile phone users. \\
	
	\item Rule-based Predictive Modeling: In general, predictive modeling uses historical data or statistics to predict a relevant outcome in the future which can be applied to any type of unknown event, regardless of when it occurred. As such the contextual behavioral rules of individual mobile phone users can be used to predict individual's behavior for a certain contextual information. Some examples of such predictions are - to predict the outgoing calls analyzing mobile phone historical call log data \cite{stefanis2014frequency,phithakkitnukoon2011behavior,plessas2017field} for smart searching in contact list, to predict incoming calls for planning and scheduling (e.g., it can be used to avoid unwanted calls and schedule time for wanted calls) \cite{phithakkitnukoon2011towards}, to predict the next mobile application that an individual is going to use for a particular contexts by analyzing individual's app usages data \cite{baeza2015predicting,kim2014conditional,zhu2012exploiting,zhu2014mobile}, to predict smart phone notification response behavior of individual users utilizing their responses to the notifications stored in the smart phone notification logs, in order to build intelligent notification management system \cite{turner2015push,kanjo2017notimind}, to assist them in their daily activities in different situations in a context-aware pervasive computing environment.
\end{itemize}

\section{Conclusion}
\label{Conclusion}
In this paper, we have discussed a number of research issues for mining contextual behavioral rules of individual mobile phone users utilizing their mobile phone data. These include the quality of smartphone data, understanding the relevancy of contexts, discretization of continuous contextual data, discovery of useful behavioral rules of individuals and their ordering, knowledge-based interactive post-mining for semantic understanding, and dynamic updating and management of rules according to their present behavior. We also discussed about the potential solution directions to overcome these issues, in order to get a complete set of behavioral rules of individual mobile phone users for using in various real-life context-aware intelligent applications discussed above. We do believe that our discussion opens a promising path for future research on mining user behaviors utilizing their mobile phone data and to build personalized rule-based intelligent systems for the end mobile phone users to intelligently assist themselves in their daily life. 

\section*{Acknowledgment}
The author would like to thank Prof. Jun Han, Swinburne University of Technology, Australia, Dr. Alan Colman, Swinburne University of Technology, Australia, and Dr. Ashad Kabir, Charles Sturt University, Australia, for their relevant discussions.

\bibliographystyle{plain}
\bibliography{bibfile/IssuesBehavMiner}

\end{document}